\documentclass[10pt,twocolumn,letterpaper]{article}

\usepackage{cvpr}
\usepackage{times}
\usepackage{epsfig}
\usepackage{graphicx}
\usepackage{amsmath}
\usepackage{amssymb}
\usepackage{enumitem}
\usepackage[normalem]{ulem}
\usepackage[ruled]{algorithm2e}
\usepackage{multirow}
\usepackage{array}
\usepackage{ctable} 
\usepackage{makecell}
\usepackage[capposition=bottom]{floatrow}
\usepackage{comment}
\DeclareMathAlphabet{\mathcal}{OMS}{cmsy}{m}{n}
\usepackage{textcomp}
\usepackage{dblfloatfix}
\usepackage{caption}
\usepackage{subcaption}
\usepackage{colortbl,xcolor}
\usepackage{booktabs}
\usepackage{mathtools}
\usepackage{textpos}

\newcommand{\name}{\textsc{VisualVoice}}

\definecolor{customgray}{rgb}{0.9, 0.9, 0.9}
\newcolumntype{g}{>{\columncolor{customgray}}c}
\newcolumntype{z}{>{\columncolor{customgray}}l}
\newcolumntype{?}[1]{!{\vrule width #1}}
\renewcommand{\paragraph}[1]{{\vspace{2mm}\noindent\textbf{#1}\,\,}}

\usepackage[pagebackref=false,breaklinks=true,letterpaper=true,colorlinks,citecolor=blue,linkcolor=blue,bookmarks=false]{hyperref}
\cvprfinalcopy 

\def\cvprPaperID{****} 

\newcommand{\speakerA}{\mathcal{A}} %
\newcommand{\speakerB}{\mathcal{B}} %

\ifcvprfinal\fi

\begin{document}

\title{\name: Audio-Visual Speech Separation with Cross-Modal Consistency}
\author{Ruohan Gao\textsuperscript{1,2}\hspace{10mm}Kristen Grauman\textsuperscript{1,3}
\\
\textsuperscript{1}The University of Texas at Austin\hspace{10mm}\hspace{10mm}\textsuperscript{2}Stanford University\hspace{10mm}\hspace{10mm}\textsuperscript{3}Facebook AI Research\\
{\tt\small rhgao@cs.stanford.edu, grauman@fb.com}
}

\maketitle
\pagestyle{empty}
\thispagestyle{empty}

\begin{abstract}

We introduce a new approach for audio-visual speech separation. Given a video, the goal is to extract the speech associated with a face in spite of simultaneous background sounds and/or other human speakers.  Whereas existing methods focus on learning the alignment between the speaker's lip movements and the sounds they generate, we propose to leverage the speaker's face appearance as an additional prior to isolate the corresponding vocal qualities they are likely to produce. Our approach jointly learns audio-visual speech separation and cross-modal speaker embeddings from unlabeled video. It yields state-of-the-art results on five benchmark datasets for audio-visual speech separation and enhancement, and generalizes well to challenging real-world videos of diverse scenarios. Our video results and code: \url{http://vision.cs.utexas.edu/projects/VisualVoice/}.
\end{abstract}
\begin{textblock*}{\textwidth}(0cm,-14.5cm)
\centering
In Proceedings of the IEEE Conference on Computer Vision and Pattern Recognition (CVPR), 2021.%
\end{textblock*}
\section{Introduction}~\label{sec:intro}
\vspace{-0.15in}

Human speech is rarely observed in a vacuum.  Amidst the noisy din of a restaurant, we concentrate to parse the words of our dining partner; watching a heated presidential debate, we disentangle the words of the candidates as they talk over one another; on a Zoom call we listen to a colleague while our children chatter and play a few yards away.  Presented with such corrupted and entangled sounds, the human perceptual system draws heavily on visual information to reduce ambiguities in the audio~\cite{rahne2007visual} and modulate attention on an active speaker in a busy environment~\cite{golumbic2013visual}. Automating this process of \emph{speech separation} has many valuable applications, including assistive technology for the hearing impaired, superhuman hearing in a wearable augmented reality device, or better transcription of spoken content in noisy in-the-wild Internet videos.

\begin{figure}
    \center
    \includegraphics[scale=0.33]{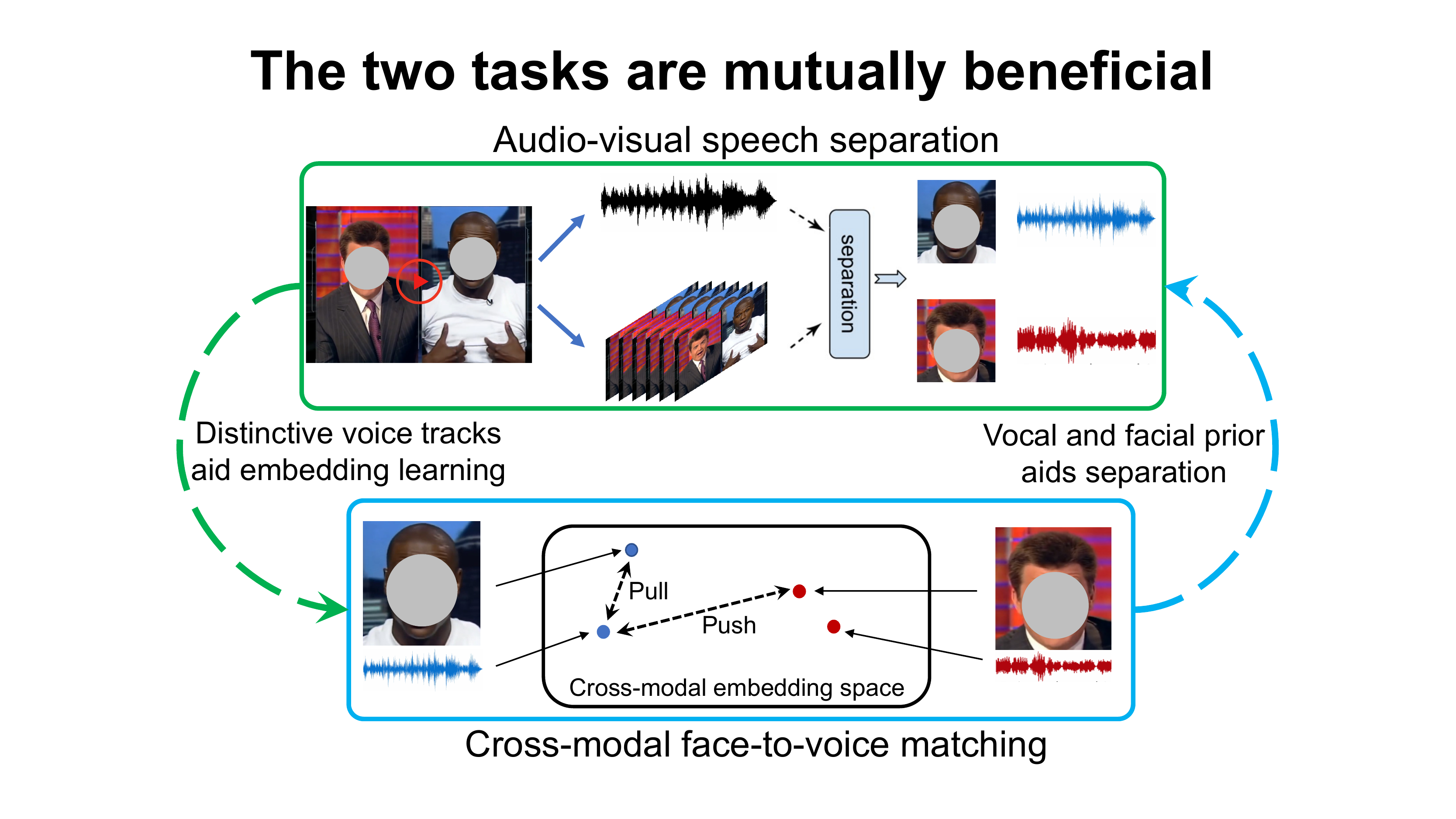}
    \caption{We propose a multi-task learning framework to jointly learn audio-visual speech separation and cross-modal face-voice embeddings. Our approach leverages the complementary cues between lip movements and cross-modal speaker embeddings for speech separation. The embeddings serve as a prior for the voice characteristics that enhances speech separation; the cleaner separated speech in turn produces more distinctive audio embeddings.}
    \label{fig:concept}
    \vspace{-0.1in}
\end{figure}

While early work in automatic speech separation relied solely on the audio stream~\cite{nakadai2002real,yilmaz2004blind,duong2010under}, recent work explores ways to leverage its close connections to the visual stream as well~\cite{gabbay2017visual, afouras2018conversation, ephrat2018looking, owens2018audio, chung2020facefilter}. By analyzing the facial motion in concert with the emitted speech, these methods steer the audio separation module towards the relevant portions of the sound that ought to be separated out from the full audio track.  For example, the mouth articulates in different shapes consistent with the phonemes produced in the audio, making it possible to mask a spectrogram for the target human speaker based on audio-visual (AV) consistency. However, solely relying on lip movements can fail when lip motion becomes unreliable, \eg, the mouth region is occluded by the microphone or the speaker turns their head away.

While AV synchronization cues are powerful, we observe that the consistency between the speaker's facial appearance and their voice is also revealing for speech separation. Intuitively, attributes like gender, age, nationality, and body weight are often visible in the face \emph{and} give a prior for what sound qualities (tone, pitch, timbre, basis of articulation) to listen for when trying to separate that person's speech from interfering sounds.  For example, female speakers often register in higher frequencies, a heavier person may exhibit a wider range of sound intensities~\cite{barsties2013body}, and an American speaker may sound more nasal.  The face-voice association, supported by cognitive science studies~\cite{bruce1986understanding}, is today often leveraged for speaker \emph{identification} given the recording of a single speaker~\cite{nagrani2018seeing,nagrani2018learnable,kim2018learning,wen2018disjoint}.  In contrast, the speech \emph{separation} problem demands discovering a cross-modal association in the presence of multiple overlapping sounds.

Our key insight is that these two tasks---cross-modal 
matching and speech separation---are mutually beneficial. The cleaner the sound separation, the more accurately an embedding can link the voice to a face; the better that embedding, the more distinctive is the prior for the voice characteristics which will in turn aid separation. We thus aim to ``visualize" the voice of a person based on how they look to better separate that voice's sound. See Figure~\ref{fig:concept}. 

To this end, we propose \name, a multi-task learning framework to jointly learn audio-visual speech separation together with cross-modal speaker embeddings. We introduce a speech separation network that takes video of a human speaker talking in the presence of other sounds (speech or otherwise) and returns the isolated sound track for just their speech. Our network relies on facial appearance, lip motion, and vocal audio to solve the separation task, augmenting the conventional ``mix-and-separate" paradigm for audio-visual separation to account for a cross-modal contrastive loss requiring the separated voice to agree with the face.  Notably, our approach requires no identity labels and no enrollment of speakers, meaning we can train and test with fully unlabeled video.

Our main contributions are as follows.  Firstly, we introduce an  audio-visual speech separation framework that leverages complementary cues from facial motion and cross-modal face-voice attributes.  Secondly, we devise a novel multi-task framework that successfully learns both separation and cross-modal embeddings in concert.  Finally, through experiments on 5 benchmark datasets, we demonstrate state-of-the-art results for audio-visual speech separation and enhancement in challenging scenarios. The embedding learned by our model additionally improves the state of the art for unsupervised cross-modal speaker verification, emphasizing the yet-unexplored synergy of the two tasks.
\section{Related Work}~\label{sec:related}
\vspace{-0.2in}

\paragraph{Audio-Only Speech Separation.} 
Sound source separation is studied extensively in auditory perception~\cite{bregman1994auditory}. Speech separation is a special case of sound source separation where the goal is to separate the speech signal of a target speaker from background interference, including non-speech noise~\cite{weninger2015speech,kumar2016speech} and/or interfering speech from other speakers~\cite{yu2017permutation,nachmani2020voice}.  While early work assumes access to multiple microphones to facilitate separation~\cite{nakadai2002real,yilmaz2004blind,duong2010under}, some methods tackle the ``blind" separation problem with monaural audio~\cite{roweis2001one,smaragdis2007supervised,spiertz2009source}, including  recent deep learning approaches~\cite{huang2014deep,wang2014training,hershey2016deep,kolbaek2017multitalker,nachmani2020voice}. Our work also targets single-channel speech separation, but unlike traditional audio-only methods, we use visual information to guide separation. 

\vspace{-0.1in}
\paragraph{Audio-Visual Source Separation.} 
Early methods that leverage audio-visual (AV) cues for source separation use techniques such as mutual information~\cite{hershey2000audio,fisher2001learning}, audio-visual independent component analysis~\cite{smaragdis2003audio,pu2017audio}, and non-negative matrix factorization~\cite{parekh2017motion,gao2018objectSounds}. Recent deep learning approaches focus on separating musical instruments~\cite{gao2019coseparation, zhao2018sound,xu2019recursive,zhao2019som,gan2020music}, speech~\cite{gabbay2017visual, afouras2018conversation, ephrat2018looking, owens2018audio, afouras2019my, chung2020facefilter}, or other sound sources from in-the-wild videos~\cite{gao2018objectSounds,tzinis2020into}. Current deep AV speech separation methods typically leverage face detection and tracking to guide the separation process~\cite{gabbay2017visual,afouras2018conversation,ephrat2018looking}, while some forgo explicit object detection and process video frames directly
~\cite{owens2018audio,Afouras20audio-visual-objects}. These methods give impressive results by learning the association between lip movements and speech.  For robustness to occluded lips, recent work incorporates a (non-visual) identity-specific voice  embedding for the audio channel~\cite{afouras2019my}. When only a profile image of the speaker is available, rather than video, learned identity embeddings extracted from a fixed pre-trained network for face images can benefit separation~\cite{chung2020facefilter}.

In contrast to prior AV methods, our model solves for source separation by incorporating both lip motion and cross-modal face-voice attributes. In particular, we propose a multi-task learning framework to jointly learn audio-visual speech separation and cross-modal speaker embeddings.
The latter helps learn separation from unlabeled video (i.e., no identity labels, no enrollment of users) by surfacing the sound properties consistent with different facial appearances, as we show in the results.

\vspace{-0.1in}
\paragraph{Cross-Modal Learning with Faces and Voices.} 
There are strong links between how a person's face looks and how their voice sounds. Leveraging this link, cross-modal learning methods explore a range of interesting tasks: face reconstruction from audio~\cite{oh2019speech2face}, talking face generation~\cite{zhou2019talking}, emotion recognition~\cite{albanie2018emotion}, speaker diarization~\cite{Nagrani20voxconverse}, speech recognition~\cite{chung2019perfect}, and speaker identification~\cite{nagrani2018seeing,nagrani2018learnable,kim2018learning,wen2018disjoint,chung2020seeing}. Unlike any of the above, our work tackles audio-visual speech separation. We jointly learn cross-modal embeddings with the goal of enhancing separation results, with the new insight that hearing voice elements consistent with a face's appearance can help disentangle speech from other overlapping sounds. 

\vspace{-0.1in}
\paragraph{Audio-Visual Learning.} 
Apart from source separation, recent inspiring work integrates both audio and visual cues on an array of other tasks including self-supervised representation learning~\cite{owens2016ambient,arandjelovic2017look,aytar2016soundnet,owens2018audio,Korbar2018cotraining,gao2020visualechoes}, localizing sounds in video frames~\cite{arandjelovic2017objects,Senocak_2018_CVPR,tian2018audio,hu2020discriminative}, generating sounds from video~\cite{owens2016visually,zhou2017visual,gao2019visualsound,morgadoNIPS18,zhou2020sep}, and action recognition~\cite{kazakos2019TBN,gao2020listentolook}. Different from all of them, our work leverages the visual cues in faces for the task of audio-visual speech separation.

\section{Approach}~\label{sec:approach}
\vspace{-0.1in}

Our goal is to perform audio-visual speech separation. We first formally define our problem (Sec.~\ref{Sec:problem_formulation}); then we present our audio-visual speech separation network (Sec.~\ref{Sec:network});
next we introduce how we learn audio-visual speech separation and cross-modal face-voice embeddings in a multi-task learning framework (Sec.~\ref{Sec:cross-modal-matching}); finally we present our training criteria and inference procedures (Sec.~\ref{Sec:training-inference}).

\subsection{Problem Formulation}~\label{Sec:problem_formulation}
Given a video $V$ with multiple speakers, we denote $x(t) = \sum_{k=1}^{K}s_k(t)$ as the observed single-channel linear mixture of the voices for these $K$ speakers, where $s_k(t)$ are time-discrete signals responsible for each speaker. Our goal in audio-visual speech separation is to separate the sound $s_k(t)$ for each speaker from $x(t)$ by leveraging the visual cues in the video. For simplicity we describe the sources as speakers throughout, but note that the mixed sound can be something other than speech, as we will demonstrate in results with speech enhancement evaluation.

To generate training examples, we follow the commonly adopted ``mix-and-separate" paradigm~\cite{ephrat2018looking,owens2018audio,afouras2018conversation,zhao2018sound,gao2019coseparation} and generate synthetic audio mixtures by mixing human speech segments. These speech segments are accompanied by the face tracks 
of the corresponding speakers, which are extracted automatically from ``in the wild" videos with background chatter, laughter, pose variation, \etc.  

Suppose we have two speech segments $s_{\speakerA_1}(t)$, $s_{\speakerA_2}(t)$ from video $V_\speakerA$ for speaker $\speakerA$, and $s_\speakerB(t)$ from video $V_\speakerB$ for speaker $\speakerB$.\footnote{No identity labels are used during training. $s_{\speakerA_1}(t)$ and $s_{\speakerA_2}(t)$ come from the same training video, so we assume they share the same identity.} Let $F_{\speakerA_1}, F_{\speakerA_2}, F_\speakerB$ denote the face tracks associated with the speech segments $s_{\speakerA_1}(t)$, $s_{\speakerA_2}(t), s_\speakerB(t)$, respectively. We create two mixture signals $x_1(t)$ and $x_2(t)$: 
\vspace{-0.05in}
\begin{equation}
    x_1(t) = s_{\speakerA_1}(t) + s_\speakerB(t),~~~
    x_2(t) = s_{\speakerA_2}(t) + s_\speakerB(t).
\end{equation}
The mixture speech signals are then transformed into complex audio spectrograms $X_1$ and $X_2$.

Our training objective is to jointly separate $s_{\speakerA_1}(t)$, $s_{\speakerA_2}(t)$ and $s_\speakerB(t)$ for face tracks $F_{\speakerA_1}$, $F_{\speakerA_2}$ and $F_{\speakerB}$ from the two mixed signals $x_1(t)$ and $x_2(t)$. In Sec.~\ref{Sec:cross-modal-matching} we present a speaker consistency loss 
that regularizes the separation process with the two mixtures. To perform separation, we predict complex ideal ratio masks (cIRM)~\cite{williamson2015complex} $M_{\speakerA_1}$, $M_{\speakerA_2}$, $M_{\speakerB_1}$ and $M_{\speakerB_2}$ to separate clean speech for the corresponding speakers from $X_1$ and $X_2$. Note that we separately predict a mask for speaker $\speakerB$ from each mixture. The predicted spectrograms for the separated speech signals are obtained by complex masking the mixture spectrograms: 
\vspace{-0.05in}
\begin{equation}
S_{\speakerA_i} = X_i \ast M_{\speakerA_i},~~S_{\speakerB_i} = X_i \ast M_{\speakerB_i},~~i\in \{1,2\},
\vspace{-0.05in}
\end{equation}
where $\ast$ indicates complex multiplication. Finally, using the inverse short-time Fourier transform (ISTFT)~\cite{griffin1984signal}, we  reconstruct the separated speech signals.

\begin{figure}[t]
    \center
    \includegraphics[scale=0.25]{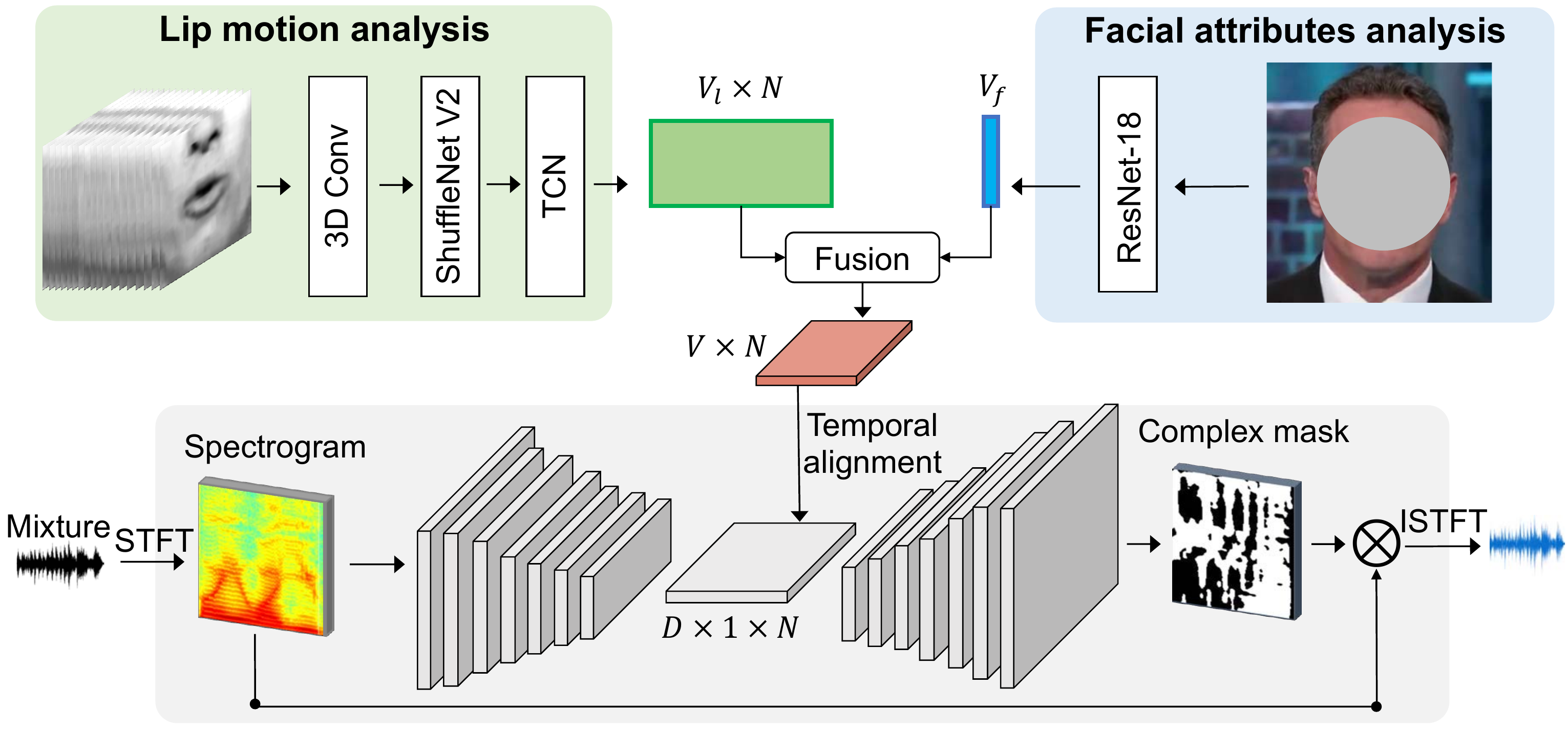}
    \caption{Our audio-visual speech separator network takes a mixed speech signal as input and analyses the lip motion and facial attributes in the face track to separate the portion of sound responsible for the corresponding speaker.}
    \label{fig:separation_network}
    \vspace{-0.05in}
\end{figure}

\subsection{Audio-Visual Speech Separator Network}~\label{Sec:network}
Next we present the architecture of our audio-visual speech separator network, which leverages the complementary visual cues of both lip motion and cross-modal facial attributes. 
Later in Sec.~\ref{Sec:cross-modal-matching} we will introduce our multi-task learning framework to learn both audio-visual speech separation and cross-modal face-voice embeddings, and describe how we jointly separate speech from $x_1(t)$ and $x_2(t)$.

We use the visual cues in the face track to guide the speech separation for each speaker. The visual stream of our network consists of two parts: a lip motion analysis network and a facial attributes analysis network (Figure~\ref{fig:separation_network}). 

Following the state-of-the-art in lip reading~\cite{martinez2020lipreading,ma2020towards}, the lip motion analysis network takes $N$ mouth regions of interest (ROIs)\footnote{The ROIs are derived from the face track through facial landmark detection and alignment to a mean reference face. See Supp. for details.} as input and it consists of a 3D convolutional layer followed by a ShuffleNet v2~\cite{ma2018shufflenet} network to extract a time-indexed sequence of feature vectors. They are then processed by a temporal convolutional network (TCN) to extract the final lip motion feature map of dimension $V_l \times N$. 

For the facial attributes analysis network, we use a ResNet-18~\cite{he2016deep} network that takes a single face image randomly sampled from the face track as input to extract a face embedding \textbf{i} of dimension $V_f$ that encodes the facial attributes of the speaker. We replicate the facial attributes feature along the time dimension to concatenate with the lip motion feature map and obtain a final visual feature of dimension $V \times N$, where $V = V_l + V_f$. 

The facial attributes feature represents an identity code whose role is to identify the space of expected frequencies or other audio properties for the speaker's voice, while the role of the lip motion is to isolate the articulated speech specific to that segment. Together they provide complementary visual cues to guide the speech separation process. 

\begin{figure*}
    \center
    \includegraphics[scale=0.49]{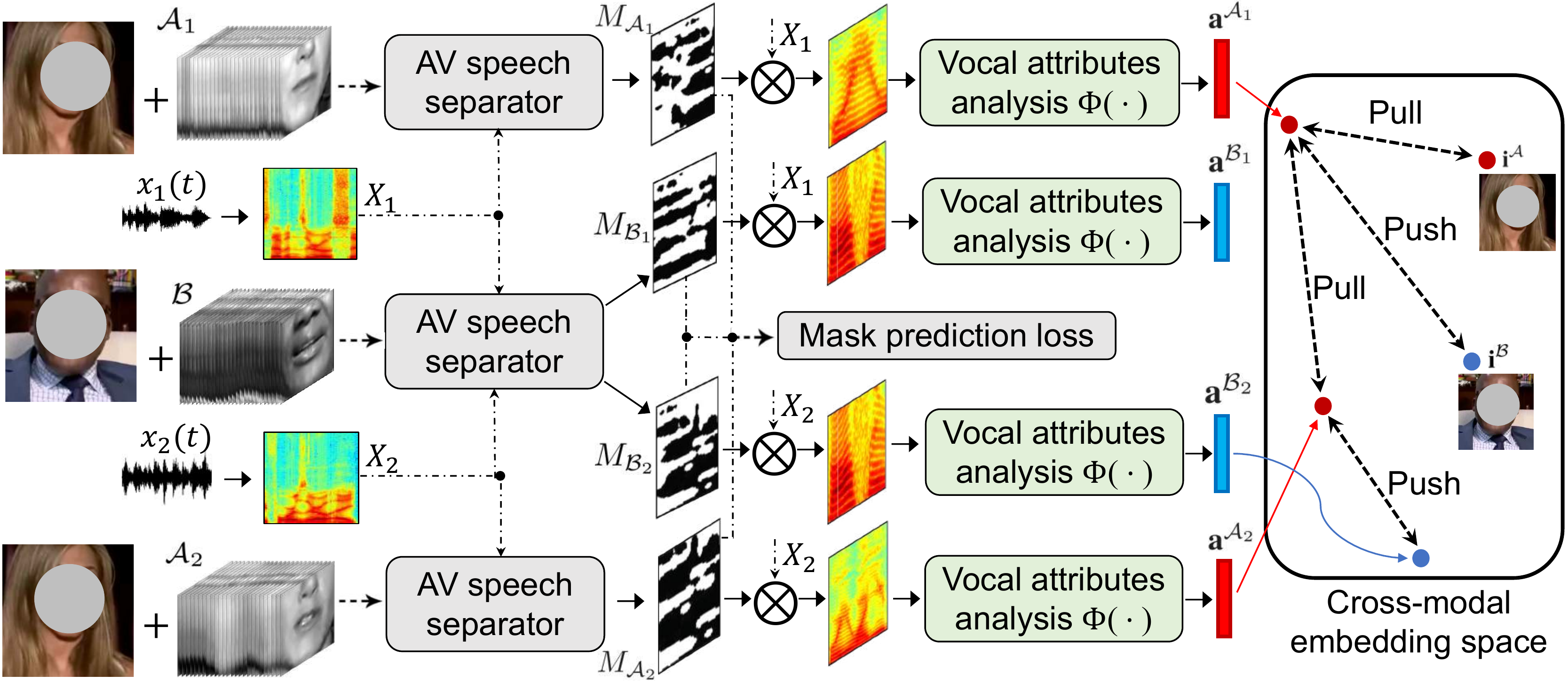}
    \caption{Our multi-task learning framework that jointly learns audio-visual speech separation and cross-modal face-voice embeddings. The network is trained by minimizing the combination of the mask prediction loss, the cross-modal matching loss, and the speaker consistency loss defined in Sec.~\ref{Sec:cross-modal-matching}.} 
    \label{fig:multi-task-network}
    \vspace{-0.1in}
\end{figure*}

On the audio side, we use a U-Net~\cite{ronneberger2015u} style network tailored to audio-visual speech separation. It consists of an encoder and a decoder network. The input to the encoder is the complex spectrogram of the mixture signal of dimension $2 \times F \times T$, where $F,T$ are the frequency and time dimensions of the spectrogram. Each time-frequency bin contains the real and imaginary part of the corresponding complex spectrogram value. The input is passed through a series of convolutional layers with frequency pooling layers in between, which reduces the frequency dimension while preserving the time dimension. In the end we obtain an audio feature map of dimension $D \times 1 \times N$, where $D$ is the channel dimension. 

We then concatenate the visual and audio features along the channel dimension to generate an audio-visual feature map of dimension $(V + D) \times 1 \times N$. The decoder takes the concatenated audio-visual feature as input.  It has symmetric structure with respect to the encoder, where the convolutional layer is replaced by an upconvolutional layer and the frequency pooling layer is replaced by a frequency upsampling layer. Finally, we use a $\mathtt{Tanh}$ layer followed by a  $\mathtt{Scaling}$ operation on the output feature map to predict a bounded complex mask of the same dimension as the input spectrogram for the speaker. 

We build an audio-visual feature map for each speaker in the mixture to separate their respective voices. Alternatively, to build a model tailored to two-speaker speech separation, we concatenate the visual features of both speakers in the mixture with the audio feature to generate an audio-visual feature map of dimension  $(2V + D) \times 1 \times N$ and simultaneously separate their voices. 
This leads to slightly better performance due to the additional context of the other speaker being provided (see Supp.~for a comparison), while a model trained with the visual feature of a single speaker can be used in the general case where the number of speakers is unknown at inference time. We use the applicable case in experiments. See Supp.~for the network details.

\subsection{Cross-Modal Matching for Separation}~\label{Sec:cross-modal-matching}
Next we introduce our multi-task learning framework that simultaneously learns AV speech separation and cross-modal face-voice embeddings. The framework includes several novel loss functions to regularize learning.

\paragraph{Mask prediction loss:} As shown in Fig.~\ref{fig:multi-task-network}, we predict complex masks $M_{\speakerA_1}$, $M_{\speakerA_2}$, $M_{\speakerB_1}$, $M_{\speakerB_2}$ to separate speech for the corresponding speakers from $X_1$ and $X_2$, respectively. We compute the following loss on the predicted complex masks:
\vspace{-0.05in}
\begin{equation}
    L_{\emph{mask-prediction}} = \sum_{i \in \{\speakerA_1, \speakerA_2, \speakerB_1, \speakerB_2\}}\|M_{i} - \mathcal{M}_{i} \|_2,
\vspace{-0.05in}
\end{equation}
where $\mathcal{M}_i$ denotes the ground-truth complex masks, which are obtained by taking the complex ratio of the spectrogram of the clean speech to the corresponding mixture speech spectrogram. This loss provides the main supervision to enforce the separation of clean speech.

\paragraph{Cross-modal matching loss:} 
To capture the desired cross-modal facial attributes to guide the separation process, we jointly learn cross-modal face-voice embeddings. The idea aligns with prior work on cross-modal matching~\cite{nagrani2018seeing,nagrani2018learnable,kim2018learning,chung2019perfect,wen2018disjoint,chung2020seeing,nagrani2020disentangled}, but here our goal is audio separation---not person identification---and rather than a single-source input, in our case the audio explicitly contains \emph{multiple} sources.

Similar to the facial attributes analysis network, we use a ResNet-18 network as the vocal attributes analysis network $\Phi(\cdot)$. We extract audio embeddings $\textbf{a}^{\speakerA_1}$, $\textbf{a}^{\speakerA_2}$, $\textbf{a}^{\speakerB_1}$, $\textbf{a}^{\speakerB_2}$ for each separated speech spectrogram:
\vspace{-0.05in}
\begin{equation}
    \textbf{a}^{\speakerA_i} = \Phi(X_i \ast M_{\speakerA_i}),~\textbf{a}^{\speakerB_i} = \Phi(X_i \ast M_{\speakerB_i}),~i \in \{1, 2\}.
\vspace{-0.05in}
\end{equation}Let $\textbf{i}^{\speakerA}$ and $\textbf{i}^{\speakerB}$ denote the face image embeddings extracted from the facial attributes analysis network for speakers $\speakerA$ and $\speakerB$, respectively. We use the following triplet loss: 
\vspace{-0.05in}
\begin{equation}
    L_{\emph{t}} (\textbf{a}, \textbf{i}^{+}, \textbf{i}^{-}) = \max \{0, D(\textbf{a}, \textbf{i}^{+}) -  D(\textbf{a}, \textbf{i}^{-}) + \mathtt{m} \},
\vspace{-0.05in}
\end{equation}
where $D(\textbf{a}, \textbf{i})$ is the cosine distance of the speech embedding and the face image embedding, and $\mathtt{m}$ represents the margin between the two distances. The cross-modal matching loss is defined as follows:
\vspace{-0.05in}
\begin{equation}
\begin{aligned}
    L_{\emph{cross-modal}} &= L_{\emph{t}} (\textbf{a}^{\speakerA_1}, \textbf{i}^{\speakerA}, \textbf{i}^{\speakerB}) + L_{\emph{t}} (\textbf{a}^{\speakerA_2}, \textbf{i}^{\speakerA}, \textbf{i}^{\speakerB}) &\\
    &+ L_{\emph{t}} (\textbf{a}^{\speakerB_1}, \textbf{i}^{\speakerB}, \textbf{i}^{\speakerA}) + L_{\emph{t}} (\textbf{a}^{\speakerB_2}, \textbf{i}^{\speakerB}, \textbf{i}^{\speakerA}).
\end{aligned}
\vspace{-0.05in}
\end{equation} 

This loss forces the network to learn cross-modal face-voice embeddings such that the distance between the embedding of the separated speech and the face embedding for the corresponding speaker should be smaller than that between the separated speech embedding and the face embedding for the other speaker, by a margin $\mathtt{m}$. It encourages the speech separation network to produce cleaner sounds so that a more accurate speech embedding can be obtained to link the voice to the face. Meanwhile, the better the face embedding, the more distinctive the facial attributes feature can be to guide the speech separation process.

\paragraph{Speaker consistency loss:} The audio segments $s_{\speakerA_1}(t)$ and $s_{\speakerA_2}(t)$ come from the same speaker from video $V_\speakerA$, so the voice characteristics of $s_{\speakerA_1}(t)$ and $s_{\speakerA_2}(t)$ should be more similar compared to $s_\speakerB(t)$. Therefore, the audio embeddings for the separated speech segments for speaker $\mathcal{A}$ should also be more similar compared to that of speaker $\mathcal{B}$. To capture this, we introduce a speaker consistency loss on the audio embeddings of the separated speech:
\begin{equation}
    L_{\emph{consistency}} = L_{\emph{t}} (\textbf{a}^{\speakerA_1}, \textbf{a}^{\speakerA_2}, \textbf{a}^{\speakerB_1}) + L_{\emph{t}} (\textbf{a}^{\speakerA_1}, \textbf{a}^{\speakerA_2}, \textbf{a}^{\speakerB_2}).
\end{equation}
This loss further regularizes the learning process by jointly separating sounds using the two mixtures.

\subsection{Training and Inference}~\label{Sec:training-inference}
The overall objective function for training is as follows:
\begin{equation}~\label{equtaion:loss}
    L =  L_{\emph{mask-prediction}} + \lambda_1 L_{\emph{cross-modal}} + \lambda_2 L_{\emph{consistency}}, 
\end{equation}
where $\lambda_1$ and $\lambda_2$ are the weight for the cross-modal matching and speaker consistency losses, respectively. During testing, we first detect faces in the video frames and extract the mouth ROIs for each speaker. For each speaker, we use the mouth ROIs and one face image (a randomly selected frame) as the visual input and predict a complex mask to separate the speech from the mixture signal. We use a sliding window approach to perform separation segment by segment for videos of arbitrary length.

Our audio-visual speech separation network is trained from scratch without using any identity labels, whereas prior methods often assume access to a pre-trained lip reading model~\cite{afouras2018conversation,afouras2019my} or a pre-trained face recognition model~\cite{ephrat2018looking} that sees millions of labeled faces. Furthermore, we do not need to pre-enroll the voice of the speakers as in~\cite{afouras2019my}. Our framework can train and test with fully unlabeled video. 
\section{Experiments}\label{sec:results}

Using a total of 6 benchmark datasets, we validate our approach for 1) audio-visual speech separation, 2) speech enhancement (Sec.~\ref{exp:av-speech-separation}), and 3) cross-modal speaker verification (Sec.~\ref{exp:av-speaker-verification}).

\begin{table*}[ht]
\begin{tabular}{c?{0.5mm}ccccc?{0.5mm}ccccc}
\multirow{2}{*}{} & \multicolumn{5}{c?{0.5mm}}{Reliable lip motion} & \multicolumn{5}{c}{Unreliable lip motion} \\ \cline{2-11} 
                  & SDR    & SIR    & SAR   & PESQ & STOI & SDR    & SIR    & SAR  & PESQ & STOI \\ \specialrule{.12em}{.1em}{.1em}
Audio-Only~\cite{yu2017permutation}   &    7.85     &     13.7       &  9.97    &  2.61  & 0.82 & 7.85     &     13.7       &  9.97    &  2.61  & 0.82\\ 
AV-Conv~\cite{afouras2018conversation}        &   8.91      &   14.8    &   11.2     &   2.73  & 0.84 & 7.23    &  11.4    &  9.98  & 2.51 & 0.80 \\ 
Ours (static face)        &   7.21      &  12.0    &   10.6     &  2.52  & 0.80 & 7.21      &  12.0    &   10.6     &  2.52  & 0.80\\ 
Ours (lip motion)   &   9.95  &   16.9      &  11.1    &   2.80     &  0.86   & 7.57 & 12.7   &  10.0    &  2.54  & 0.81  \\ 
Ours        &  \textbf{10.2}      &    \textbf{17.2}    &     \textbf{11.3}   &     \textbf{2.83}     & \textbf{0.87} & \textbf{8.53}   &  \textbf{14.3}    &  \textbf{10.4} & \textbf{2.64} & \textbf{0.84} \\
\end{tabular}

\caption{Audio-visual speech separation results on the VoxCeleb2 dataset. We show the performance separately for testing examples where the lip motion is reliable (left) or unreliable (right). See text for details. Higher is better for all metrics. }
\vspace{-0.1in}
\label{Table:av-speech-separation}
\end{table*}

\subsection{Datasets}
\textbf{VoxCeleb2~\cite{Nagrani18}:} This dataset contains over 1 million utterances with the associated face tracks extracted from YouTube videos, with 5,994 identities in the training set and 118 identities in the test set. We hold out two videos for each identity in the training set as our seen-heard test set, and we use 59 identities in the original test set as our validation set and the other 59 identities as our unseen-unheard test set. Note that we make use of the identity labels only for the purpose of making these evaluation splits. During testing, we randomly mix two test clips from different speakers to create the synthetic mixture. This ensures the ground-truth of the separated speech is known for quantitative evaluation, following standard practice~\cite{afouras2018conversation,ephrat2018looking}. We randomly sample 2,000 test parings each from the seen-heard and unseen-unheard test sets. For speech enhancement experiments, we additionally mix the speech mixture with non-speech audios from AudioSet~\cite{gemmeke2017audio} as background noise during both training and testing. The types of noise include music, laughter, crying, engine, wind, \etc. See Supp.~for details and video examples.

\textbf{Mandarin~\cite{hou2018audio}}, \textbf{TCD-TIMIT~\cite{harte2015tcd}}, \textbf{CUAVE~\cite{patterson2002cuave}, LRS2\footnote{All experiments on LRS2 were conducted at UT Austin.}~\cite{afouras2018deep}}: We evaluate on these four standard benchmark datasets to compare our model with a series of state-of-the-art audio-visual speech separation and enhancement methods in Sec.~\ref{sec:sota}. See Supp. for details.

\textbf{VoxCeleb1~\cite{Nagrani17}:} This dataset contains over 100,000 utterances for 1,251 celebrities extracted from YouTube videos. We evaluate on this dataset for cross-modal speaker verification in Sec.~\ref{exp:av-speaker-verification}. We use the same train/val/test split  as in~\cite{nagrani2018learnable} to compare with their reported results.

We are conscious of the risk of biases in data-driven methods for human understanding and have taken measures to mitigate them; see Supp.~for broader impact discussion.

\subsection{Implementation Details}  

Our AV speech separation framework is implemented in PyTorch. For all experiments, we sub-sample the audio at 16kHz, and the input speech segment is 2.55s long. STFT is computed using a Hann window length of 400 with a hop size of 160 and FFT window size of 512. The complex spectrogram $X$ is of dimension $2 \times 257 \times 256$. The input to the lip motion analysis network is $N = 64$ mouth regions of interest (ROIs) of size of $88 \times 88$, and the input to the face attributes analysis network is a face image of size $224 \times 224$. The lip motion feature is of dimension $V_l \times N$ with $V_l = 512$, $N = 64$. The dimension for both the face and voice embeddings is 128. The entire network is trained using an Adam optimizer with weight decay of 0.0001, batchsize of 128, and starting learning rate set to $1 \times 10^{-4}$. $\lambda_1$ and $\lambda_2$ are both set to 0.01 in Eq.~\ref{equtaion:loss}. The loss terms are not normalized for scale, so the absolute values of the loss weights do not directly indicate their impact on learning. The margin $\mathtt{m}$ is set to 0.5 for the triplet loss. See Supp.~for details of the network architecture and other optimization hyperparameters.

\subsection{Results on Audio-Visual Speech Separation}~\label{exp:av-speech-separation}
We first evaluate on audio-visual speech separation and compare to a series of state-of-the-art methods~\cite{ephrat2018looking,afouras2018conversation,Afouras20audio-visual-objects,chung2020facefilter,gabbay2017visual,pu2017audio,casanovas2010blind,hou2018audio} and multiple baselines: \\
\vspace{-0.2in}
\begin{itemize}
    \item \textbf{Audio-Only}: This baseline uses the same architecture as our method except that no visual feature is used to guide the separation process. We use the permutation invariant loss (PIT)~\cite{yu2017permutation} to train the network.
    \vspace{-0.1in}
    \item \textbf{Ours (lip motion)}: An ablation of our method where only the lip motion analysis network is used to guide the separation process.
    \vspace{-0.1in}
    \item \textbf{Ours (static face)}: An ablation of our method where only the facial attributes analysis network is used to guide the separation process.
    \vspace{-0.1in}
    \item \textbf{AV-Conv~\cite{afouras2018conversation}}: A state-of-the-art audio-visual speech separation method that predicts the magnitude and phase of the spectrogram separately through two subnetworks. Because the authors' code is available, we can use it for extensive experiments trained and evaluated on the same data as our method.
    \vspace{-0.1in}
    \item \textbf{Ephrat~\etal~\cite{ephrat2018looking}, Afouras~\etal~\cite{Afouras20audio-visual-objects}, Chung~\etal~\cite{chung2020facefilter}, Gabbay~\etal~\cite{gabbay2017visual}, Hou~\etal~\cite{hou2018audio}, Casanovas~\etal~\cite{casanovas2010blind}, Pu~\etal~\cite{pu2017audio}}: We directly quote results from~\cite{ephrat2018looking,Afouras20audio-visual-objects,chung2020facefilter} to compare to a series of prior state-of-the-art methods on standard benchmarks in Sec.~\ref{sec:sota}.
\end{itemize}

We evaluate the speech separation results using a series of standard metrics including Signal-to-Distortion Ratio (SDR), Signal-to-Interference Ratio (SIR), and Signal-to-Artifacts Ratio (SAR) from the mir eval library~\cite{raffel2014mir_eval}. We also evaluate using two speech-specific metrics: Perceptual Evaluation of Speech Quality (PESQ)~\cite{rix2001perceptual}, which measures the overall perceptual quality of the separated speech and Short-Time Objective Intelligibility (STOI)~\cite{taal2011algorithm}, which is correlated with the intelligibility of the signal.

\begin{table*}[ht]
\begin{tabular}{c?{0.5mm}ccccc?{0.5mm}ccccc}
\multirow{2}{*}{} & \multicolumn{5}{c?{0.5mm}}{Reliable lip motion} & \multicolumn{5}{c}{Unreliable lip motion} \\ \cline{2-11} 
                  & SDR    & SIR    & SAR   & PESQ & STOI & SDR    & SIR    & SAR  & PESQ & STOI \\ \specialrule{.12em}{.1em}{.1em}
Audio-Only~\cite{yu2017permutation}   &    3.56     &     10.9      &  5.71    &  2.00  & 0.66 & 3.56     &     10.9      &  5.71    &  2.00  & 0.66 \\ 
AV-Conv~\cite{afouras2018conversation}        &   5.32      &   11.9   &   7.52     &   2.20   & 0.71 & 3.99    &  9.43    &  6.92  & 2.02 & 0.67 \\ 
Ours (static face)        &   3.48      &  8.43    &   6.91     &  1.96   & 0.68 & 3.48      &  8.43    &   6.91     &  1.96   & 0.68 \\ 
Ours (lip motion)        &   6.31      &  13.3   &   7.72     &  2.32   & 0.76 & 4.21    &  9.78    &  6.85  & 2.03 & 0.69 \\ 
Ours        &  \textbf{6.55}      &    \textbf{13.7}    &  \textbf{7.84}   &    \textbf{2.34}     & \textbf{0.77} & \textbf{4.95}    &  \textbf{11.0}   &  \textbf{7.02}  & \textbf{2.12} & \textbf{0.72} \\
\end{tabular}

\caption{Audio-visual speech enhancement results on the VoxCeleb2 dataset with audios from AudioSet used as non-speech background noise. Higher is better for all metrics.}
\vspace{-0.1in}
\label{Table:av-speech-enhancement}
\end{table*}

\subsubsection{Quantitative Results}
Table~\ref{Table:av-speech-separation} shows the speech separation results on the VoxCeleb2 dataset. We use the visual features of both speakers as input to guide the separation and simultaneously separate their voices. We present results separately for scenarios where the lip motion is reliable and unreliable. For the reliable case, we use the original mouth ROIs extracted automatically from the face tracks; for the unreliable case, we randomly shift the mouth ROI sequences in time by up to 1s and occlude the lip region for up to 1s per segment during both training and testing. These corruptions represent typical video artifacts (e.g., buffering lag) and mouth occlusions. Table~\ref{Table:av-speech-enhancement} shows the speech enhancement results. The setting is the same as Table~\ref{Table:av-speech-separation} except that the mixture contains additional background sounds (e.g., laughter, car engine, wind, \etc) sampled from AudioSet. The visual feature of only the target speaker is used to guide the separation for speech enhancement experiments.

Tables~\ref{Table:av-speech-separation} and~\ref{Table:av-speech-enhancement} show that in both scenarios, our method achieves the best separation results.~It outperforms AV-Conv~\cite{afouras2018conversation} by a good margin. The audio-only baseline benefits from our architecture design, and it has decent performance, though note that unlike AV methods, it cannot assign the separated speech to the corresponding speaker. We evaluate both possible matchings and report its best results (to the baseline's advantage). The ablations show that separation with our model is possible purely using one static face image, but it can be difficult especially when the facial attributes alone are not reliable or distinctive enough to guide separation (see Supp.). Lip motion is directly correlated with the speech content and is much more informative for speech separation when reliable. However, the performance of the lip motion-based model significantly drops when the lip motion is unreliable, as often the case in real-world videos. Our \name~approach combines the complementary cues in both the lip motion and the face-voice embedding learned with cross-modal consistency, and thus is less vulnerable to unreliable lip motion. See Supp.~for an ablation for the different loss terms.

\begin{table}
\begin{subtable}{\linewidth}\centering
{\resizebox{1\linewidth}{!}{
\begin{tabular}{zccccc}
\toprule
     & Gabbay~\etal~\cite{gabbay2017visual} & Hou~\etal~\cite{hou2018audio} & Ephrat~\etal~\cite{ephrat2018looking} & Ours
     \\ \hline
PESQ  & 2.25  & 2.42 &  2.50 & \textbf{2.51} \\ 
STOI  & --    & 0.66 &  0.71 & \textbf{0.75} \\ 
SDR   &  --   & 2.80 &  6.10 & \textbf{6.69}   \\ 
\bottomrule
\end{tabular}
}}
\caption{Results on Mandarin dataset.}
\vspace{0.05in}
\label{subtable:mandarin}
\end{subtable}
\begin{subtable}{\linewidth}\centering
{\resizebox{1\linewidth}{!}{
\begin{tabular}{zcccc}
\toprule
     & Gabbay~\etal~\cite{gabbay2017visual} & Ephrat~\etal~\cite{ephrat2018looking} & Ours
     \\ \hline
SDR  &   0.40  &  4.10 &  \textbf{10.9} \\
PESQ  &   2.03  & 2.42  &  \textbf{2.91} \\
\bottomrule
\end{tabular}
}}
\caption{Results on TCD-TIMIT dataset.}
\vspace{0.05in}
\label{subtable:tcdtimit}
\end{subtable}
\begin{subtable}{\linewidth}\centering
{\resizebox{1\linewidth}{!}{
\begin{tabular}{zccccc}
\toprule
     & Casanovas~\etal~\cite{casanovas2010blind} & Pu~\etal~\cite{pu2017audio} &  Ephrat~\etal~\cite{ephrat2018looking} & Ours \\
     \hline
SDR  &   7.0  &  6.2 &  12.6 & \textbf{13.3}\\
\bottomrule
\end{tabular}
}}
\caption{Results on CUAVE dataset.}
\vspace{0.05in}
\label{subtable:cuave}
\end{subtable}
\begin{subtable}{\linewidth}\centering
{\resizebox{1\linewidth}{!}{
\begin{tabular}{zccc}
\toprule
     & Afouras~\etal~\cite{afouras2018conversation} & Afouras~\etal~\cite{Afouras20audio-visual-objects} & Ours \\
     \hline
SDR &  11.3 &  10.8  &  \textbf{11.8} \\
PESQ &  3.0 &   3.0  &  \textbf{3.0} \\
\bottomrule
\end{tabular}
}}
\caption{Results on LRS2 dataset.}
\vspace{0.05in}
\label{subtable:lrs2}
\end{subtable}
\begin{subtable}{\linewidth}\centering
{\resizebox{1\linewidth}{!}{
\begin{tabular}{zcccc}
\toprule
     & Chung~\etal~\cite{chung2020facefilter} & Ours (static face) & Ours \\
     \hline
SDR  &   2.53  & 7.21 & \textbf{10.2}\\
\bottomrule
\end{tabular}
}}
\caption{Results on VoxCeleb2 dataset.}
\vspace{0.05in}
\label{subtable:facefilter}
\end{subtable}
\caption{Comparing to prior state-of-the-art methods on audio-visual speech separation and enhancement. Baseline results are quoted from~\cite{ephrat2018looking,Afouras20audio-visual-objects,chung2020facefilter}.
\vspace{-0.2in}
}
\label{Table:compare_to_sota}
\vspace{-0.05in}
\end{table}

\vspace{-0.1in}
\subsubsection{Comparison to State-of-the-Art Methods}~\label{sec:sota}
\vspace{-0.15in}

Table~\ref{Table:compare_to_sota} compares our method to a series of state-of-the-art methods on AV speech separation and enhancement.  We use the same evaluation protocols and the same metrics. Our approach improves the state-of-the-art on each of the five datasets.

Whereas Tables~\ref{Table:av-speech-separation} and~\ref{Table:av-speech-enhancement} use the exact same training sources for all methods, here we rely on the authors' reported results~\cite{ephrat2018looking,Afouras20audio-visual-objects,chung2020facefilter} in the literature to make comparisons, which draw on different sources. In Table~\ref{subtable:mandarin}-\ref{subtable:cuave}, we evaluate on the Mandarin, TCD-TIMIT and CUAVE datasets using our speaker-independent model trained on VoxCeleb2 to test the cross-dataset generalization capability of our model. Note that this setting is similar to~\cite{ephrat2018looking}, where they also use a speaker-independent model trained on AV-Speech to test on these datasets. In comparison, the other prior methods require training a speaker-dependent model for each speaker in the test dataset. Our model significantly outperforms these methods, despite never seeing the speakers during training. In Table~\ref{subtable:lrs2}, we train and test on the LRS2 dataset following~\cite{Afouras20audio-visual-objects}. Our method consistently outperforms all these prior methods. Notably, in Table~\ref{subtable:facefilter}, our ablated static face model trained with cross-modal consistency significantly improves the prior static image-based model FaceFilter~\cite{chung2020facefilter} by 4.68 in SDR. This shows that the cross-modal speaker embeddings learned through our \textsc{VisualVoice} framework can provide sufficient cues for separation, even without using any information on lip movements. This is important for a wide range of scenarios (\eg, online social network platforms) where videos containing lip motion are absent, but a user's profile image is available to use for separation.

\vspace{-0.1in}
\subsubsection{Qualitative Results}

\textbf{Real-World Speech Separation.} To further test our method's success on real-world videos with mixed speech, we run our model on a variety of 
test videos in various challenging scenarios including presidential debates, zoom calls, interviews, noisy restaurants, \etc. Note that these videos lack ground-truth, but can be manually checked for quality as shown in the Supp.~video.

\vspace{0.05in}
\noindent \textbf{Best/Worst Performing Pairs.} 
We illustrate the best and worst performing pairs for speech separation using synthetic pairs for our static face model in the Supp. Pairs that perform best tend to be very different in terms of facial attributes like gender, age, and nationality. Speech separation can be hard if the two mixed identities are visually similar or the facial attributes are hard to obtain from only a static face image due to occlusion or irregular pose.

To further understand when the cross-modal face-voice embeddings help the most, we also compare the per-pair performance of our model with only lip motion and our full model. The pairs with the largest improvement from the cross-modal face-voice embeddings tend to be those that either have very different facial appearances or whose lip motion cues are difficult to extract (\eg, non-frontal views). See Supp.~for examples.

\subsection{Learned Cross-Modal Embeddings}~\label{exp:av-speaker-verification}
Our results thus far show how the cross-modal embedding learning enhances speech separation, our primary goal. As a byproduct of our AV speech separation framework, cross-modal embedding learning may also benefit from our model's joint learning. Thus we next evaluate the cross-modal verification task, in which the system must decide if a given face and voice belong to the same person.

To compare with prior cross-modal learning work, we train and evaluate on the VoxCeleb1 dataset and compare with the following baselines: 1) \textbf{Learnable-Pins~\cite{nagrani2018learnable}}: A state-of-the-art cross-modal embedding learning method. We directly quote their reported results and follow the same evaluation protocols and data splits to compare with our method; 2) \textbf{Random}: Embeddings extracted from a randomly initialized network of the same architecture as our method; 3) \textbf{Ours (single-task)}: Our cross-modal embedding network without jointly training for speech separation.

Table~\ref{Table:cross-modal-verification} shows the results. We use standard metrics for verification.
Our cross-modal embedding network alone compares favorably with~\cite{nagrani2018learnable} on seen-heard speakers and generalizes much better to unseen-unheard speakers. When trained with speech separation in a multi-task setting, our method achieves large gains, demonstrating that our idea to jointly train for these two tasks is beneficial to learn more reliable cross-modal face-voice embeddings.  
Table~\ref{Table:cross-modal-verification} provides an apples-to-apples comparison, whereas methods following~\cite{nagrani2018learnable} evaluate under several different protocols.  Other new loss function designs and the extra supervision used in the other prior work are orthogonal to our idea and could also augment our model.

\begin{figure}[t]
    \center
    \includegraphics[scale=0.255]{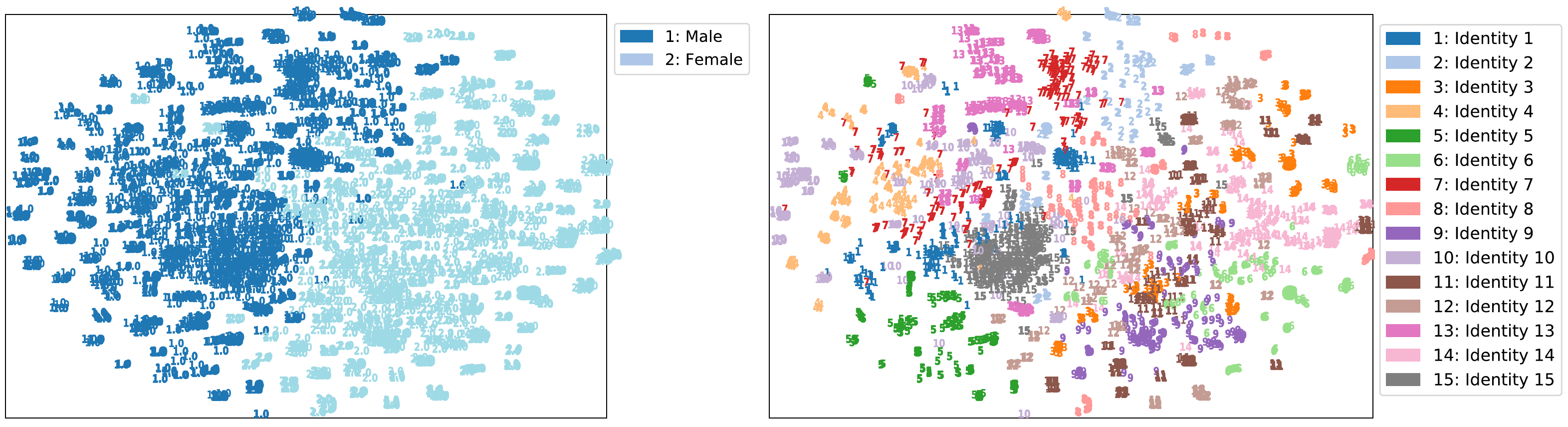}
    \caption{Our learned cross-modal embeddings of voices for 15 speakers from the VoxCeleb1 test set visualized with t-SNE. The two figures are color coded with gender and identity, respectively.}
    \label{fig:tsne}
    \vspace*{-0.05in}
\end{figure}

To \emph{visualize} that our \name~framework has indeed learned useful cross-modal face-voice embeddings, Figure~\ref{fig:tsne} shows the t-SNE~\cite{maaten2008visualizing} embeddings of the voices for 15 random speakers from the VoxCeleb1 test set. The embeddings are extracted from our vocal attributes analysis network jointly trained with speech separation. The two sub-figures are color-coded with gender and identity, respectively. Our method's learned voice embeddings tend to cluster speakers of the same cross-modal attributes together despite having access to no identity labels and no attribute labels during training.

\begin{table}[t]
{\resizebox{1\linewidth}{!}{
\begin{tabular}{c?{0.5mm}cc?{0.5mm}cc}
\multirow{2}{*}{} & \multicolumn{2}{c?{0.5mm}}{Seen-Heard} & \multicolumn{2}{c}{Unseen-Unheard} \\ \cline{2-5} 
                  & AUC~$\uparrow$    & EER~$\downarrow$   & AUC~$\uparrow$     &  EER~$\downarrow$  \\ \specialrule{.12em}{.1em}{.1em}
Random   &    50.8     &     49.6       &  49.7    &  50.1   \\ 
Learnable Pins~\cite{nagrani2018learnable}        &   73.8      &   34.1    &   63.5     &    39.2       \\ 
Ours (single-task)       &  75.0      &    32.2    &     72.4   &    34.7        \\
Ours      &  \textbf{84.9}      &    \textbf{23.6}    &     \textbf{74.2}   &     \textbf{32.3}         \\
\end{tabular}
}
\caption{Cross-modal verification results on the VoxCeleb1 dataset. $\downarrow$ lower better, $\uparrow$ higher better.}
\label{Table:cross-modal-verification}
}
\vspace{-0.05in}
\end{table}

\section{Conclusion}

We presented an audio-visual speech separation framework that simultaneously learns cross-modal speaker embeddings and speech separation in a multi-task setting. Our \textsc{VisualVoice} approach exploits the complementary cues between the lip motion and cross-modal facial attributes. It achieves state-of-the-art results on audio-visual speech separation and generalizes well to challenging real-world videos. 
Our design for the cross-modal matching and speaker consistency losses is not restricted to the speech separation task, and can be potentially useful for other audio-visual applications, such as learning intermediate features for speaker identification and sound source localization. As future work, we plan to explicitly model the fine-grained cross-modal attributes of faces and voices, and leverage them to further enhance speech separation.

\vspace{-0.05in}
\begin{small}
\paragraph{Acknowledgements:} Thanks to Ariel Ephrat, Triantafyllos Afouras, and Soo-Whan Chung for help with experiments setup and to Lorenzo Torresani, Laurens van der Maaten, and Sagnik Majumder for feedback on paper drafts. RG is supported by a Google PhD Fellowship and a Adobe Research Fellowship.  UT Austin is supported in part by NSF IIS-1514118 and the IFML NSF AI Institute.
\end{small}



{\small
\bibliographystyle{ieee}
\bibliography{ref_RG.bib}
}

\end{document}